\documentclass[conference]{IEEEtran}
\IEEEoverridecommandlockouts
\usepackage{cite}
\usepackage[ruled,vlined]{algorithm2e}
\usepackage{amsmath,amssymb,amsfonts}
\usepackage{algorithmic}
\usepackage{multirow}
\usepackage{graphicx}
\usepackage{textcomp}
\usepackage{xcolor}
\usepackage{caption}  
\usepackage{subcaption} 
\usepackage{float}
\def\BibTeX{{\rm B\kern-.05em{\sc i\kern-.025em b}\kern-.08em
    T\kern-.1667em\lower.7ex\hbox{E}\kern-.125emX}}

\newcommand{\methodname}{{\tt{DAG-AFL}}}
\begin{document}

\title{\methodname{}: Directed Acyclic Graph-based Asynchronous Federated Learning

\author{
    \IEEEauthorblockN{Shuaipeng Zhang$^{1,2}$, Lanju Kong$^{1,2*}$, Yixin Zhang$^{1,2}$, Wei He$^{1,2}$,Yongqing Zheng$^{1,2}$, Han Yu$^{3}$, Lizhen Cui$^{1,2*}$ \thanks{\IEEEauthorrefmark{0*} Lanju Kong and Lizhen Cui are the corresponding authors.}\IEEEauthorrefmark{0}}
    \IEEEauthorblockA{$^1$ School of Software, Shandong University, Jinan, China}
    \IEEEauthorblockA{$^2$ Joint SDU-NTU Centre for Artificial Intelligence Research (C-FAIR), Shandong University, China}
    \IEEEauthorblockA{$^3$  College of Computing and Data Science, Nanyang Technological University, Singapore}
    \IEEEauthorblockA{\{202320848,yixinzhang\}@mail.sdu.edu.cn,\{klj, hewei, clz\}@sdu.edu.cn, zhengyongqing@dareway.com.cn,han.yu@ntu.edu.sg}}
}

\maketitle

\begin{abstract}
Due to the distributed nature of federated learning (FL), the vulnerability of the global model and the need for coordination among many client devices pose significant challenges. As a promising decentralized, scalable and secure solution, blockchain-based FL methods have attracted widespread attention in recent years. However, traditional consensus mechanisms designed for Proof of Work (PoW) similar to blockchain incur substantial resource consumption and compromise the efficiency of FL, particularly when participating devices are wireless and resource-limited. To address asynchronous client participation and data heterogeneity in FL, while limiting the additional resource overhead introduced by blockchain, we propose the \underline{D}irected \underline{A}cyclic \underline{G}raph-based \underline{A}synchronous \underline{F}ederated \underline{L}earning (\methodname{}) framework. We develop a tip selection algorithm that considers temporal freshness, node reachability and model accuracy, with a DAG-based trusted verification strategy. Extensive experiments on 3 benchmarking datasets against eight state-of-the-art approaches demonstrate that \methodname{} significantly improves training efficiency and model accuracy by 22.7\% and 6.5\% on average, respectively.
\end{abstract}

\begin{IEEEkeywords}
Federated Learning, Blockchain, DAG, Tip Selection
\end{IEEEkeywords}

\section{Introduction}
\label{sec:intro}
With the rise of multimedia applications, large datasets have become essential for training effective machine learning (ML) models \cite{chen2022noise}. However, directly accessing distributed datasets, such as mobile device usage data, might compromise user privacy. Regulations like GDPR thus restrict data sharing. Federated learning (FL) \cite{ren2025advances,fan2025ten} has emerged as a promising collaborative ML paradigm enabling efficient model training without direct access to sensitive local data.

Despite the significant potential of FL in ensuring data privacy, its practical deployment remains challenging. In particular, device asynchrony and data heterogeneity have emerged as critical factors influencing FL performance. 
Device asynchrony \cite{ren2025advances} encompasses variations in nodes’ computational power, communication conditions, storage capacities, battery levels, data volumes, and training durations. Such variability complicates system coordination and impairs training efficiency.  For instance, in centralized synchronous FL systems, such as the one proposed by Google in 2017 \cite{mcmahan2017communication}, each node must await the completion of all others’ tasks before advancing to the next training round.
Furthermore, low computing capacity or unexpected device disconnections can result in overall training failure. 
Additionally, the lack of visibility into data and operational processes among FL nodes significantly complicates efforts to address data heterogeneity \cite{li2022data}. Variations in data distributions across nodes can reduce system efficiency, degrade model accuracy, and increase the vulnerability of the aggregated model. Consequently, it is crucial to implement effective mechanisms to tackle data heterogeneity in FL. To this end, some studies focus on asynchronous FL to address device asynchrony issues while mitigating the impact of data heterogeneity.

Blockchain offers a natural fit for FL's asynchronous and secure environment \cite{zhu2023blockchain}, as miners can verify parameters, maintain normal operations, and filter abnormal nodes. Yet, most blockchain-based FL approaches still follow the synchronous FL model, introducing pseudo-heterogeneity and requiring miner interconnections, which can affect performance (latency, convergence, accuracy). Moreover, reliance on Proof-of-Work (PoW) \cite{nakamoto2008bitcoin} consumes significant resources, potentially undermining performance in resource-limited settings.

The blockchain based on directed acyclic graph (DAG) \cite{wang2023sok} represents an emerging paradigm, it is more suitable for implementing decentralized FL in applications with stringent timeliness requirements compared to traditional blockchains. Unlike traditional blockchains, DAG-based systems allow clients to upload transactions during idle periods without waiting for miners. This approach reduces resource consumption and improves overall efficiency. Furthermore, the DAG structure maintains clients' direct communication with other clients, making it an ideal architecture for FL \cite{cao2021toward}. 

In this paper, we propose the \textbf{\underline{DAG}}-based \textbf{\underline{A}}synchronous \textbf{\underline{F}}ederating \textbf{\underline{L}}earning method (\methodname{}) customized for edge devices, to address the issues of device asynchrony and data heterogeneity. We design a DAG-based tip selection method to improve FL model accuracy. For added security, we design a DAG verification strategy to prevent task publishers from tampering with the overall DAG structure. Extensive experimental results show that our method achieves more efficient federated learning training without compromising accuracy. Compared to eight state-of-the-art methods over three benchmarking datasets, \methodname{} improves training efficiency by 22.7\% and model accuracy by 6.5\% on average.


\section{RELATED WORK}

\subsection{Federated Learning}

Federated learning (FL) has gained prominence as a privacy-preserving distributed learning paradigm. Although FedAvg \cite{mcmahan2017communication} is a foundational method, it struggles with data heterogeneity and device asynchrony, impairing model performance and slowing convergence. To address these challenges, researchers have explored asynchronous FL algorithms. For instance, FedAsync \cite{xie2019asynchronous} uses asynchronous communication with adaptive learning ratios but risks inconsistent parameters and delayed global convergence. Semi-asynchronous approaches, such as FedAT \cite{chai2021fedat}, combine synchronous and asynchronous training based on client response latencies to mitigate lag. Additionally, methods like FedHiSyn \cite{li2022fedhisyn} cluster clients by data distribution and computational capacity, improving update efficiency and communication.

\subsection{Blockchain-based federated learning}

In recent years, integrating federated learning with blockchain has made substantial progress. 
BlockFL\cite{kim2019blockchained} uses blockchain to upload all locally updated models, it allows clients to download and aggregate new global models. The participants are given rewards for training local models, validating local updates, and creating new blocks. BFLC \cite{li2020blockchain} introduces an enhanced FL framework with a committee-based consensus mechanism, to mitigate malicious attacks and reduce computational overhead. To further improve efficiency, ScaleSFL\cite{madill2022scalesfl} adopts a sharding approach that integrates both shard-level and main-chain consensus.
\subsection{DAG-based Federated Learning}
Unlike traditional blockchains, DAG-based ledgers employ a nonlinear data structure that enables parallel transaction processing, increasing throughput and reducing confirmation times. Nodes validate and approve earlier transactions before broadcasting their own, and unapproved transactions with suitable staleness are called tips \cite{popov2018tangle}.
In the Internet of Vehicles (IoV), multiple DAGs store training models for vehicle groups, while Roadside Units (RSUs) aggregate and upload the global model to a traditional blockchain \cite{lu2020blockchain}. Although \cite{lu2020blockchain} uses cumulative weights to evaluate DAG-based model data, it does not address non-IID data distributions. Similarly, \cite{yuan2021chainsfl} uploads training results to a local blockchain and then records the aggregated model on a DAG ledger, focusing on security but overlooking non-IID data challenges.

\section{System design}
\subsection{Architecture Design of DAG-AFL}
Our architecture defines two primary roles:

\textbf{Task publisher}: Initializes and oversees the federated learning process. It provides an initial global model on the DAG, monitors node status, and issues a termination command once the target accuracy is reached. Unlike traditional centralized FL, the publisher does not train models but focuses on real-time coordination.

\begin{figure}[htbp]
\centerline{\includegraphics[width=1\linewidth]{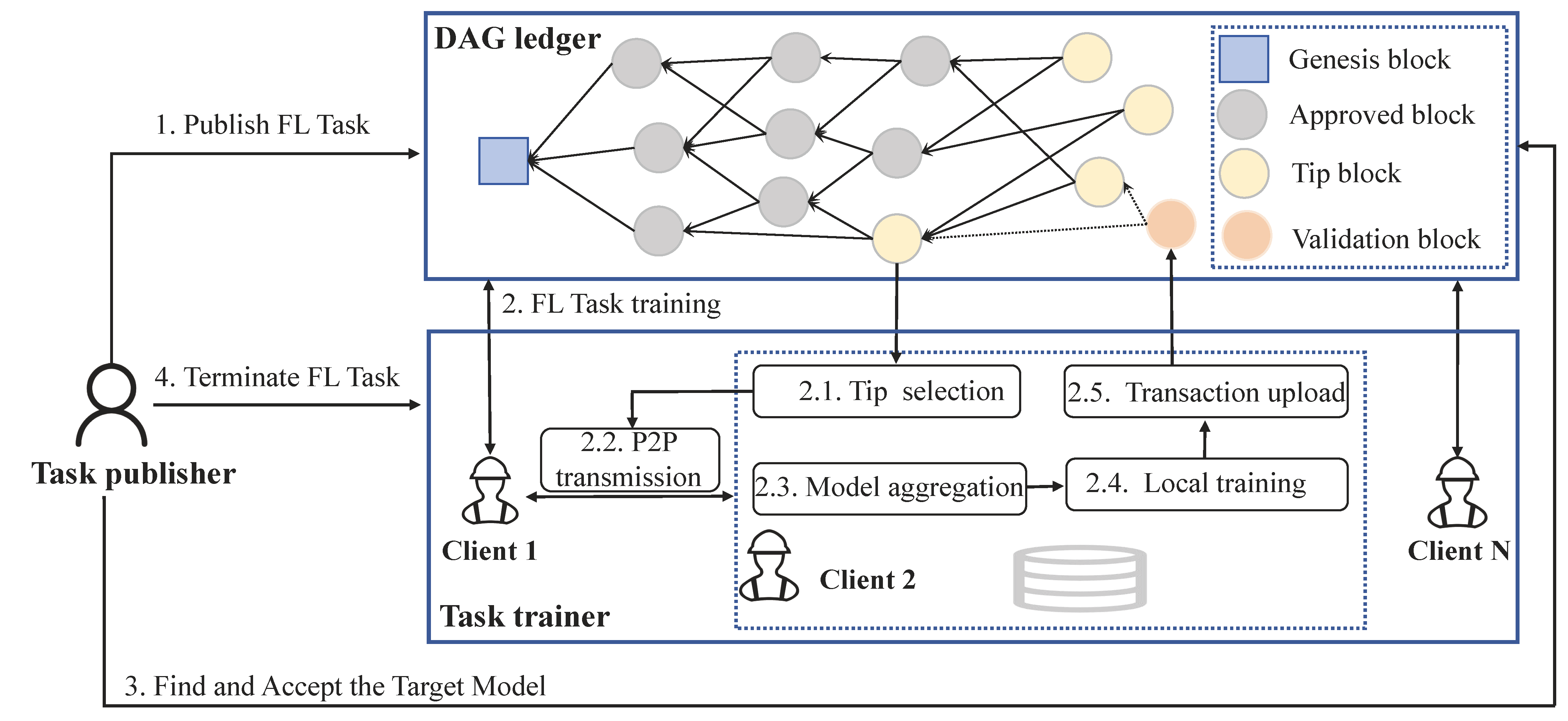}}
\vspace{-5pt}
\caption{DAG-AFL Architecture Overview.}
\label{fig}
\end{figure}
\vspace{-8pt}

\textbf{Task trainer}: Manages local training and helps maintain the DAG. It selects suitable tips to retrieve metadata, metadata includes critical information such as node IDs and model accuracy but does not contain the model itself (details in Section III-C). Using this metadata, the trainer identifies relevant client nodes for collaboration. It then acquires the required models through a peer-to-peer (P2P) protocol. These models are aggregated into a new training model, which the trainer updates on the DAG along with its metadata. This process repeats iteratively until the model achieves the desired accuracy or the set number of iterations is completed.

The workflow proceeds as follows: the task publisher initializes FL tasks and selects trainers. Authorized trainers pull the initial model from DAG, train it locally, and upload updates. Nodes use tip selection algorithms to identify peer models via P2P, aggregate them into a new global model for local training, then iteratively update the DAG until meeting iteration limits or accuracy thresholds.

\subsection{Tip selection}
Tip selection is a crucial step in DAG-based FL, as it directly influences local model training accuracy. Efficient tip selection substantially improves overall training efficiency and final model accuracy. The tip selection algorithm’s primary objective is to identify which tips a client should approve when issuing the next transaction. 
We will introduce an efficient and precise method for selecting tips, focusing on three key dimensions: tip freshness, tip reachability, and model accuracy. 

\subsubsection{Tip freshness}

In a DAG, the existence time of tips varies, and tips close to the current time generally exhibit greater freshness. Since the DAG tends to evolve toward improved model accuracy, higher freshness often corresponds to more accurate model states. 

The global iteration epochs of different trainers exert varying influences on the updates of tips. When the global iteration epochs of the selected tip significantly differ from those of the current trainer, it signifies a notable disparity in the number of updates between the two, consequently leading to differences in their training content and effectiveness. In other words, tips with smaller epoch gaps share more similar training content and progress, thus providing more consistent contributions to model refinement. To this end, the global iteration epoch difference of tips is regarded as a key factor for freshness, denoted as \textit{Tipc}. To quantify its effect,  \textit{Tipc} is normalized to the range [0, 1] using an exponential function. The calculation formula is as follows:
\begin{equation}
Tipc(k)={e}^{-|{T}_{cur}-{T}_{tip(k)}|},k\in \{1,2,...,m\} \label{eq}
\end{equation}

Among them, $T_{cur}$ represents the global iteration epoch of the current trainer $cur$, $T_{tip(k)}$ denotes the global iteration epoch of the $k$-th tip, and $m$ indicates the total number of tips at the current time. For current trainer $cur$, 
 the freshness of tips $k$ is calculated as follows:
\begin{multline}
Freshness(k)=\\
Tipc(k)\cdotp \frac{1}{1+\alpha \cdotp (current\_time-Time(tip(k))}
\end{multline}

Among them, $\alpha$ is the decay factor controlling the rate of freshness decay over time, with larger $\alpha$ values increasing sensitivity to time differences. In a DAG, a tip’s dwell time, calculated as $current\_time-Time(tip(k))$, indicates its temporal distance from the current time. Longer dwell times reduce freshness, diminishing the tip's contribution to model accuracy. Tip freshness enables real-time evaluation and timely model updates.


\subsubsection{Tip Reachability}
\begin{figure}[htbp]
\centerline{\includegraphics[width=0.45\linewidth]{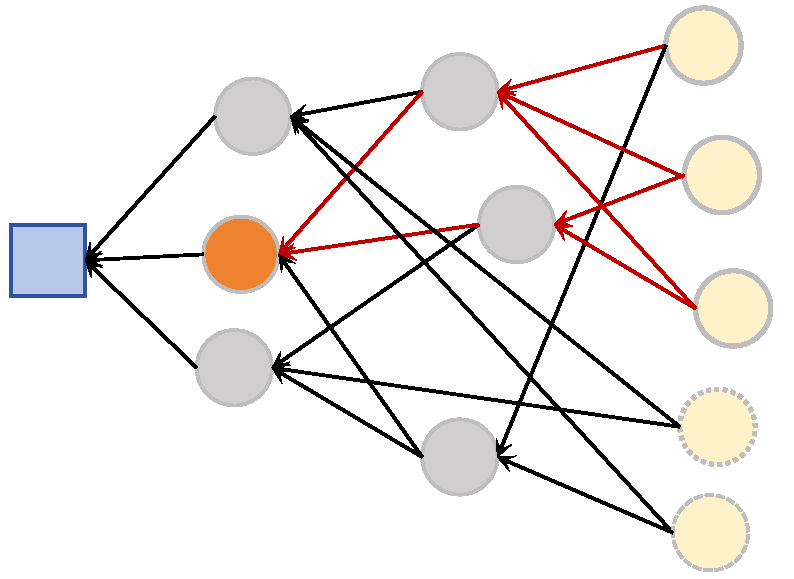}}
\caption{Example of Tip Reachability.}
\label{fig2}
\end{figure}

The reachability of a tip indicates that it has directly or indirectly integrated results from the previous aggregation model in the current iteration. This suggests a similar data distribution between the tip and the querying node. Conversely, unreachable tips may exhibit greater uncertainty in data distribution, necessitating their differentiation. In essence, tip reachability reflects the similarity of data distribution across different clients. In the DAG (see Fig.~\ref{fig2}), the orange node represents the latest uploaded model of a client, solid light yellow nodes denote reachable tips from this client, while dashed light yellow nodes represent unreachable tips. Red arrows indicate the reachable paths.

\vspace{-8pt}
\vspace{-8pt}

To determine a tip's reachability, we apply a Breadth-First Search (BFS) algorithm to identify all indirectly linked tips, as demonstrated in Algorithm\ref{algorithm1}. The overall time complexity of BFS is $O(V+E)$, where $V$ represents the number of nodes in the graph and $E$ denotes the number of edges in the graph. Importantly, rather than starting from the genesis block, the DAG search initiates from the client’s most recent node.

\begin{algorithm}[ht]
\caption{Finding Reachable and Unreachable Tips in a DAG via BFS}
\label{algorithm1}
\KwIn{DAG, the latest uploaded model of a client $startNode$}
\KwOut{ReachableTips (reachable tips from $startNode$), UnreachableTips (all other tips with in-degree = 0)}

Initialize an empty queue $Q$ and an empty set $visited$\;
Initialize empty sets: $ReachableTips$ and $AllTips$\;
Add all nodes with in-degree = 0 to $AllTips$\;
$Q.\text{add}(startNode)$\;
$visited.\text{add}(startNode)$\;

\While{$Q$ is not empty}{
    $node = Q.\text{dequeue()}$\;
    \If{$inDegree(node) = 0$}{
        $ReachableTips.\text{add}(node)$\;
    }
    \ForEach{$neighbor$ in $Adj(node)$}{
        \If{$neighbor \notin visited$}{
            $Q.\text{add}(neighbor)$\;
            $visited.\text{add}(neighbor)$\;
        }
    }
}

$UnreachableTips = AllTips - ReachableTips$\;
\Return $ReachableTips, UnreachableTips$\;
\end{algorithm}
\vspace{-2pt}

However, relying solely on reachable nodes for training data may cause the model to converge to a local optimum. To avoid this issue, we adopt balances the proportion of reachable and unreachable tips. Specifically, the algorithm reduces excessive dependence on similar data distributions, enhances model generalization, and avoids premature convergence.

\subsubsection{Model accuracy}
It is a critical metric for evaluation, and variations in accuracy directly affect the selection and application of models. Existing tip selection methods often verify the accuracy of all tips in each iteration, which is resource-intensive. To improve efficiency, we employ a feature distribution similarity approach to choose tips, reducing computational overhead while maintaining selection quality.

To efficiently identify tips suited to a given client, we assign a feature signature to each tip, inspired by PFA\cite{liu2021pfa}. PFA leverages the sparsity of neural networks to represent a client’s original data and group clients by similar data distributions. Building on this idea, we introduce a feature extraction technique based on local training. Due to the high uniqueness of feature maps, they serve as 'signatures' that distinguish various data distributions. By using these feature signatures, the algorithm can quickly identify nodes with analogous data distributions, thereby significantly reducing redundant accuracy computations.

Given a set of clients, each client $K_i$ holds a private dataset $D_i$ with $N_i$ samples. Each client trains a local model on its own dataset and selects several kernel functions from an intermediate layer. During forward propagation, the client extracts a feature matrix from the output of the chosen kernel function for a given input $x_h$. The $k$-th input's signature is then computed as follows.
\begin{equation}
sig_k(x_h)=\frac{zero(F_k(x_h))}{H\times W} \label{eq}
\end{equation}

In this context, the $zero(\cdotp )$ function counts the number of zero elements in the matrix, and $F_k(x_h)$ denotes the output feature matrix of the selected kernel function. For the dataset $D_i$, the signature of the $k$-th kernel function is obtained by averaging the output features across all samples:
\begin{equation}
sig_k(D_i)=\frac{1}{N}\displaystyle\sum_{h=1}^{N_i}sig_k(x_h) \label{eq}
\end{equation}

During local training, each client $K_i$ extracts a signature vector $S_i = {sig_1(D_i), sig_2(D_i), ..., sig_n(D_i)} $ and stores it on the blockchain, reflecting the data distribution of that client. The similarity between clients is then measured by computing the cosine similarity of their signature vectors, where $S_i$ and $S_j$ denote the signature vectors of clients $K_i$ and $K_j$, respectively.
\begin{equation}
sim(K_i,K_j)=cos(S_i,S_j)=\frac{S_i\cdotp S_j}{||S_i||\cdotp ||S_j||} \label{eq}
\end{equation}

We implement a smart contract to compute and store client similarity. This contract maintains a similarity matrix that records similarity scores for each training round, enabling subsequent queries. These similarity values guide the selection of accuracy criteria in each model iteration. As the client model updates with every iteration, its feature parameters change accordingly. Therefore, each client must upload its feature values as signatures for the corresponding tip.

Consider a scenario where the client selects $N$ ($N<M$) tips from a total of $M$ tips, comprised of $N_1 = \lambda N$ reachable tips (from $M_1$ currently available)  and $N_2 = (1-\lambda)N$ unreachable tips (from the remaining $M-M_1$), with $\lambda$ as a constant coefficient. For the set of reachable tips, the client directly evaluates model accuracy and selects the top $N_1$ tips. For unreachable tips, it first samples based on data distribution similarity. From the similarity matrix, it chooses $p$ ($p \leq M-M_1$ and $p \geq N_2$) tips that exhibit the highest similarity to the current node. These $p$ tips are validated using the current client's test set, ranked by accuracy, and the top $N_2$ are selected as the final unreachable tips.

The $N$ selected tips are then aggregated by averaging their models. The formula for computing the aggregated model of the $k$-th client in the $t$-th round is given by:
\begin{equation}
{w}_{k}^{t}=\frac{1}{N}(\displaystyle\sum_{i=1}^{\lambda N}{w}_{i}^{t-1}+\displaystyle\sum_{j=1}^{(1-\lambda)N}{w}_{j}^{t-1}) \label{eq}
\end{equation}

Use the aggregated model as the local model and initiate a new training epoch. After training concludes, retain the resulting model as the client’s global model and extract its feature signatures for the next training phase. At this point, the client only needs to upload the metadata: $<ClientId, Signature, ModelAccuracy, CurrentEpoch, \\ValidationNodeId>$.

\subsection{Trustworthy verification of DAG}
The task publisher retains the entire DAG for trainers to use during tip selection and model validation. To ensure the publisher cannot tamper with the DAG by task, a verification path is introduced. Trainers store only specific validation paths from the DAG. By checking these paths, they verify the authenticity of submitted data and ensure that the publisher has not maliciously altered DAG contents.

During training, the trainers validate the consistency of the publisher's data by comparing the stored hash values with the actual data state. Consistent hash values indicate unaltered data, whereas inconsistencies suggest potential tampering.

After deriving node reachability from Algorithm 1, each tip's hash is generated in two parts. The first part is the hash of the two referenced tips, serving as the block header, and the second is the hash of the uploaded metadata, serving as the block’s content body. The calculation process is as follows:
\begin{align}
\text{Hash}(tip_{i})
&= \bigl\{ H_{1}, H_{2}, \nonumber\\
&\quad \text{hash}(ClientId \mid Signature \mid \nonumber\\
&\quad\quad ModelAccuracy \mid CurrentEpoch \mid \nonumber\\
&\quad\quad ValidationNodeId) \bigr\}\label{eq:hash_tipi}
\end{align}

Here,$H_{1}$ and $H_{2}$ denote the hash values of the two tips referenced by the current block, functioning as block header pointers to ensure an ordered and traceable blockchain structure. The $hash(\cdot)$ function processes the client ID, signature, model accuracy, current epoch, and validation node ID to generate a block body digest, thereby ensuring the integrity and immutability of federated learning model updates.

During verification, the client only needs to retain the hash of the current tip. By backtracking through the referenced blocks, the client can confirm the trustworthiness of the entire chain, reflecting the intrinsic resistance of the blockchain to tampering and strong auditability.

\section{Experimental Analysis}

\subsection{Experiment Preparation}\label{AA}
\textbf{Environment.} All experiments were conducted on a PC equipped with a GeForce RTX 4080 GPU, an Intel Core i9-13900k 3.00GHz CPU and 64GB memory. We implmented \methodname{} in Python with PyTorch. 
\begin{table}[]
\caption{Federated learning datasets}
\label{tab1}
\begin{tabular}{|c|c|c|c|}
\hline
\textbf{Datasets} & \textbf{Samples} & \textbf{Classes} & \textbf{Task descriptions}    \\ \hline
Mnist             & 70000            & 10               & Handwritten digit recognition \\ \hline
CIFAR-10          & 60000            & 10               & Image classification          \\ \hline
CIFAR-100         & 60000            & 100              & Image classification          \\ \hline
\end{tabular}
\end{table}

\textbf{Datasets.}  Table \ref{tab1} summarizes the three benchmark datasets(MNIST, CIFAR-10, and CIFAR-100). Each dataset is randomly partitioned into training, validation, and testing sets at an 8:1:1 ratio. We conduct experiments under both IID and non-IID data distributions. In the IID setting, the dataset is evenly split among clients, ensuring uniform data sizes. In the non-IID setting, we simulate heterogeneous distributions using a Dirichlet distribution with parameters 0.1 and 0.05 (denoted as $\beta$), where smaller values of $\beta$ indicate greater data heterogeneity and size deviation.

\textbf{FL Competitors}. We compare our method with two baselines with independent (each client trains locally) and centralized (no data privacy, using the entire dataset for a global model), as well as Fedavg\cite{mcmahan2017communication}, FedAsync\cite{xie2019asynchronous}, CSAFL\cite{zhang2021csafl}, FedAT\cite{chai2021fedat}, and FedHisyn\cite{li2022fedhisyn}. Both FedAvg and FedHiSyn adopt synchronous FL, FedAsync is asynchronous, and CSAFL and FedAT are semi-asynchronous.

\textbf{Blockchain Competitors}. BlockFL\cite{kim2019blockchained}, BFLC\cite{li2020blockchain}, and ScaleSFL\cite{madill2022scalesfl} integrate synchronous FL with open-source implementations. Since BlockFL and BFLC rely on FedAvg, we select ScaleSFL for performance comparisons. DAG-FL is from~\cite{cao2021toward}.

\textbf{Implementation Details}. We use VGG16~\cite{simonyan2014very} as the backbone for MNIST, CIFAR-10, and CIFAR-100. Hyperparameters are tuned via grid search. The convolution kernel size is set to 3×3. We choose 10 clients and a maximum of 200 global iterations. Early stopping relies on validation-set average accuracy with a patience of 5 rounds. Each client trains locally for 5 epochs per round, and the learning rate is 0.01. By default, each trainer selects two tips, sets the parameter $\lambda=0.5$, and employs a freshness decay factor $\alpha=0.1$.
\vspace{-0.3cm}
\subsection{performance comparison in Federated Learning}
\textbf{Accuracy}. Table \ref{tab2} reports the average accuracy, defined as the proportion of correctly predicted samples in the test set. Three key observations emerge:
(1) Across most configurations, \methodname{} consistently ranks among the top two methods in terms of average accuracy, achieving 99.56\% on MNIST IID, outperforming DAG-FL's 98.45\%;
(2) The centralized setting achieves the highest accuracy (99.64\% on MNIST), while the independent setting has the lowest (24.14\% on CIFAR-100, $\beta=0.05$), representing the theoretical upper and lower bounds, respectively;
(3) Asynchronous methods show lower accuracy than synchronous ones, e.g., FedAsync reaches 82.84\% on CIFAR-10 ($\beta=0.1$), below ScaleSFL's 85.39\%. This occurs because asynchronous approaches trade some accuracy to achieve higher processing efficiency.

\begin{table*}[htbp]
\centering
\caption{Comparison of average accuracy by different methods on the MNIST, CIFAR-10, and CIFAR-100 datasets.}
\label{tab2}
\begin{tabular}{|c|ccc|ccc|ccc|}
\hline
\multirow{2}{*}{\textbf{Accuracy}} & \multicolumn{3}{c|}{\textbf{MNIST}} & \multicolumn{3}{c|}{\textbf{CIFAR-10}} & \multicolumn{3}{c|}{\textbf{CIFAR-100}} \\ \cline{2-10} 
                                   & \multicolumn{1}{c|}{\textbf{IID}} & \multicolumn{1}{c|}{\textbf{$\beta$=0.1}} & \textbf{$\beta$=0.05} & \multicolumn{1}{c|}{\textbf{IID}} & \multicolumn{1}{c|}{\textbf{$\beta$=0.1}} & \textbf{$\beta$=0.05} & \multicolumn{1}{c|}{\textbf{IID}} & \multicolumn{1}{c|}{\textbf{$\beta$=0.1}} & \textbf{$\beta$=0.05} \\ \hline
\textbf{Centralized}               & \multicolumn{1}{c|}{99.64} & \multicolumn{1}{c|}{99.64} & 99.64 & \multicolumn{1}{c|}{90.23} & \multicolumn{1}{c|}{90.23} & 90.23 & \multicolumn{1}{c|}{67.42} & \multicolumn{1}{c|}{67.42} & 67.42 \\ 
\textbf{Independent}               & \multicolumn{1}{c|}{88.62} & \multicolumn{1}{c|}{80.22} & 76.84 & \multicolumn{1}{c|}{77.21} & \multicolumn{1}{c|}{47.10} & 41.23 & \multicolumn{1}{c|}{46.22} & \multicolumn{1}{c|}{30.98} & 24.14 \\ \hline
\textbf{FedAvg}                    & \multicolumn{1}{c|}{99.55} & \multicolumn{1}{c|}{\textit{99.54}} & \textit{99.52} & \multicolumn{1}{c|}{88.26} & \multicolumn{1}{c|}{83.55} & 78.01 & \multicolumn{1}{c|}{61.24} & \multicolumn{1}{c|}{55.02} & 51.36 \\ 
\textbf{FedHisyn}                  & \multicolumn{1}{c|}{99.52} & \multicolumn{1}{c|}{99.35} & 99.21 & \multicolumn{1}{c|}{\textit{89.73}} & \multicolumn{1}{c|}{\textbf{85.91}} & 68.82 & \multicolumn{1}{c|}{\textit{66.79}} & \multicolumn{1}{c|}{\textit{62.27}} & \textbf{58.73} \\ 
\textbf{ScaleSFL}                  & \multicolumn{1}{c|}{\textbf{99.58}} & \multicolumn{1}{c|}{\textbf{99.55}} & 99.49 & \multicolumn{1}{c|}{89.22} & \multicolumn{1}{c|}{85.39} & \textbf{82.95} & \multicolumn{1}{c|}{65.57} & \multicolumn{1}{c|}{61.58} & \textit{58.66} \\ \hline
\textbf{FedAsync}                  & \multicolumn{1}{c|}{99.55} & \multicolumn{1}{c|}{99.53} & 99.47 & \multicolumn{1}{c|}{89.54} & \multicolumn{1}{c|}{82.84} & \textit{80.54} & \multicolumn{1}{c|}{65.48} & \multicolumn{1}{c|}{57.87} & 52.23 \\ 
\textbf{CSAFL}                     & \multicolumn{1}{c|}{99.53} & \multicolumn{1}{c|}{99.51} & 99.50 & \multicolumn{1}{c|}{89.58} & \multicolumn{1}{c|}{71.74} & 67.94 & \multicolumn{1}{c|}{66.60} & \multicolumn{1}{c|}{57.36} & 52.27 \\ 
\textbf{DAG-FL}                    & \multicolumn{1}{c|}{98.45} & \multicolumn{1}{c|}{98.30} & 98.25 & \multicolumn{1}{c|}{87.29} & \multicolumn{1}{c|}{81.24} & 70.21 & \multicolumn{1}{c|}{62.31} & \multicolumn{1}{c|}{56.23} & 53.14 \\ \hline
\textbf{\methodname{} (ours)}             & \multicolumn{1}{c|}{\textit{99.56}} & \multicolumn{1}{c|}{\textit{99.54}} & \textbf{99.53} & \multicolumn{1}{c|}{\textbf{89.79}} & \multicolumn{1}{c|}{\textit{85.86}} & 80.34 & \multicolumn{1}{c|}{\textbf{66.80}} & \multicolumn{1}{c|}{\textbf{62.53}} & 57.89 \\ \hline
\end{tabular}
\end{table*}

\textbf{Training time}. Table \ref{tab3} compares the average training time required for global convergence:
(1) FedAsync achieves the shortest training time, such as 1,154 seconds on MNIST IID and 1,892 seconds on CIFAR-10 IID, but at the cost of significantly lower accuracy; 
(2) Asynchronous methods, such as FedAsync and \methodname{}, generally outperform synchronous and semi-asynchronous approaches in average training time, as they avoid delays caused by waiting. For instance, \methodname{} trains in 1,302 seconds on MNIST IID, significantly faster than FedAvg's 1,684 seconds and FedHisync's 6,176 seconds; 
(3) Except for FedAsync, \methodname{} generally offers superior performance. \methodname{} balances accuracy and training time effectively. On CIFAR-100 non-IID with $\beta=0.05$, it takes 3,316 seconds, outperforming DAG-FL (3,418 seconds) and CSAFL (3,746 seconds), making it suitable for time-sensitive, heterogeneous environments.

\begin{table*}[]
\centering
\caption{Comparison of average training time by different methods on the MNIST, CIFAR-10, and CIFAR-100 datasets.}
\label{tab3}
\begin{tabular}{|c|ccc|ccc|ccc|}
\hline
\multirow{2}{*}{\textbf{Time(second)}} & \multicolumn{3}{c|}{\textbf{Mnist}}                                                     & \multicolumn{3}{c|}{\textbf{Cifar-10}}                                                  & \multicolumn{3}{c|}{\textbf{Cifar-100}}                                                 \\ \cline{2-10} 
                                       & \multicolumn{1}{c|}{\textbf{IID}}  & \multicolumn{1}{c|}{\textbf{$\beta$=0.1}}  & \textbf{$\beta$=0.05} & \multicolumn{1}{c|}{\textbf{IID}}  & \multicolumn{1}{c|}{\textbf{$\beta$=0.1}}  & \textbf{$\beta$=0.05} & \multicolumn{1}{c|}{\textbf{IID}}  & \multicolumn{1}{c|}{\textbf{$\beta$=0.1}}  & \textbf{$\beta$=0.05} \\ \hline
\textbf{Centralized}                   & \multicolumn{1}{c|}{1493}          & \multicolumn{1}{c|}{1493}          & 1493          & \multicolumn{1}{c|}{2150}          & \multicolumn{1}{c|}{2150}          & 2150          & \multicolumn{1}{c|}{2145}          & \multicolumn{1}{c|}{2145}          & 2145          \\
\textbf{Independent}                   & \multicolumn{1}{c|}{1235}          & \multicolumn{1}{c|}{1231}          & 1165          & \multicolumn{1}{c|}{1632}          & \multicolumn{1}{c|}{1758}          & 1926          & \multicolumn{1}{c|}{1689}          & \multicolumn{1}{c|}{1869}          & 1848          \\ \hline
\textbf{Fedavg}                        & \multicolumn{1}{c|}{1684}          & \multicolumn{1}{c|}{1594}          & \textit{1354}          & \multicolumn{1}{c|}{3535}          & \multicolumn{1}{c|}{4242}          & 4054          & \multicolumn{1}{c|}{3854}          & \multicolumn{1}{c|}{4569}          & 4234          \\
\textbf{FedHisyn}                      & \multicolumn{1}{c|}{6176}          & \multicolumn{1}{c|}{5096}          & 7245          & \multicolumn{1}{c|}{15128}         & \multicolumn{1}{c|}{23969}         & 10547         & \multicolumn{1}{c|}{24623}         & \multicolumn{1}{c|}{24303}         & 27531         \\
\textbf{ScaleSFL}                      & \multicolumn{1}{c|}{3268}          & \multicolumn{1}{c|}{3356}          & 2539          & \multicolumn{1}{c|}{3631}          & \multicolumn{1}{c|}{6269}          & 6379          & \multicolumn{1}{c|}{24623}         & \multicolumn{1}{c|}{24303}         & 27531         \\ \hline
\textbf{FedAsync}                      & \multicolumn{1}{c|}{\textbf{1154}} & \multicolumn{1}{c|}{\textbf{1035}} & \textbf{1069} & \multicolumn{1}{c|}{\textbf{1892}} & \multicolumn{1}{c|}{\textbf{1822}} & \textbf{1752} & \multicolumn{1}{c|}{\textbf{3259}} & \multicolumn{1}{c|}{\textbf{3120}} & \textbf{3452} \\
\textbf{CSAFL}                         & \multicolumn{1}{c|}{1384}          & \multicolumn{1}{c|}{1222}          & 1954          & \multicolumn{1}{c|}{2047}          & \multicolumn{1}{c|}{3311}          & 3086          & \multicolumn{1}{c|}{4249}          & \multicolumn{1}{c|}{4213}          & 3746          \\
\textbf{DAG-FL}                        & \multicolumn{1}{c|}{1426}          & \multicolumn{1}{c|}{1332}          & 1658          & \multicolumn{1}{c|}{2321}          & \multicolumn{1}{c|}{2846}          & 4221          & \multicolumn{1}{c|}{3451}          & \multicolumn{1}{c|}{3384}          & 3418          \\ \hline
\textbf{\methodname{}(ours)}                 & \multicolumn{1}{c|}{\textit{1302}}          & \multicolumn{1}{c|}{\textit{1215}}          & 1594          & \multicolumn{1}{c|}{\textit{2036}}        & \multicolumn{1}{c|}{\textit{2754}}          & \textit{2848}         & \multicolumn{1}{c|}{\textit{3362}}          & \multicolumn{1}{c|}{\textit{3224}}          & \textit{3316}          \\ \hline
\end{tabular}
\end{table*}
\vspace{-8pt}

\subsection{performance comparison in blockchain}
In the blockchain environment, we evaluated the throughput (transactions per second, TPS) and latency (average confirmation time) of various blockchain-based FL systems. By adjusting the number of clients, we examined core functionalities such as uploading updated models and querying the latest global models. For DAG-based FL, the primary query cost arises from the tip selection process.

\vspace{-10pt}
\begin{figure}[htbp]
    \centering
    \begin{subfigure}[b]{0.22\textwidth}
        \centering
        \includegraphics[width=\textwidth]{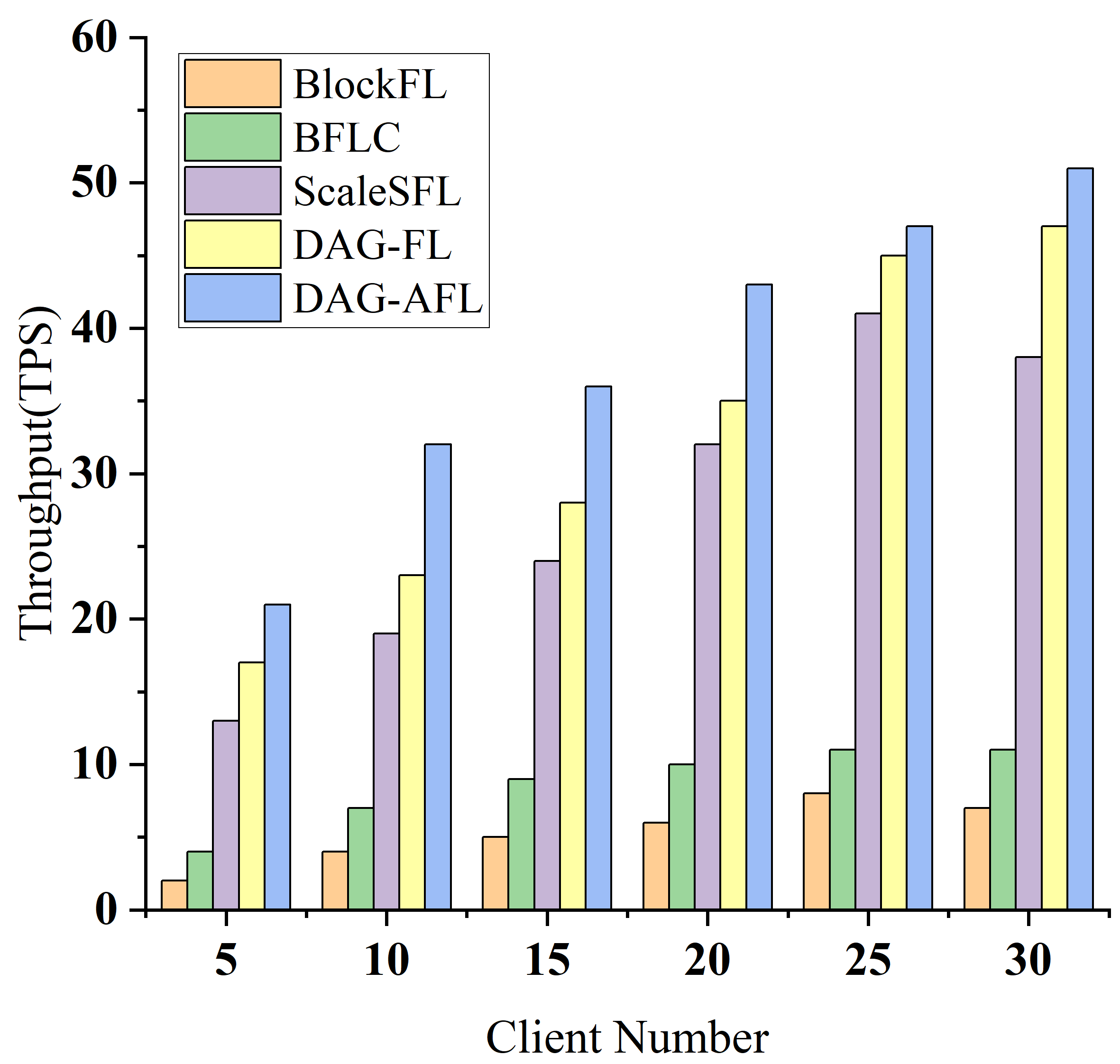}
         \caption{}
        \label{fig:subfig1}
    \end{subfigure}
    \hfill
    \begin{subfigure}[b]{0.22\textwidth}
        \centering
        \includegraphics[width=\textwidth]{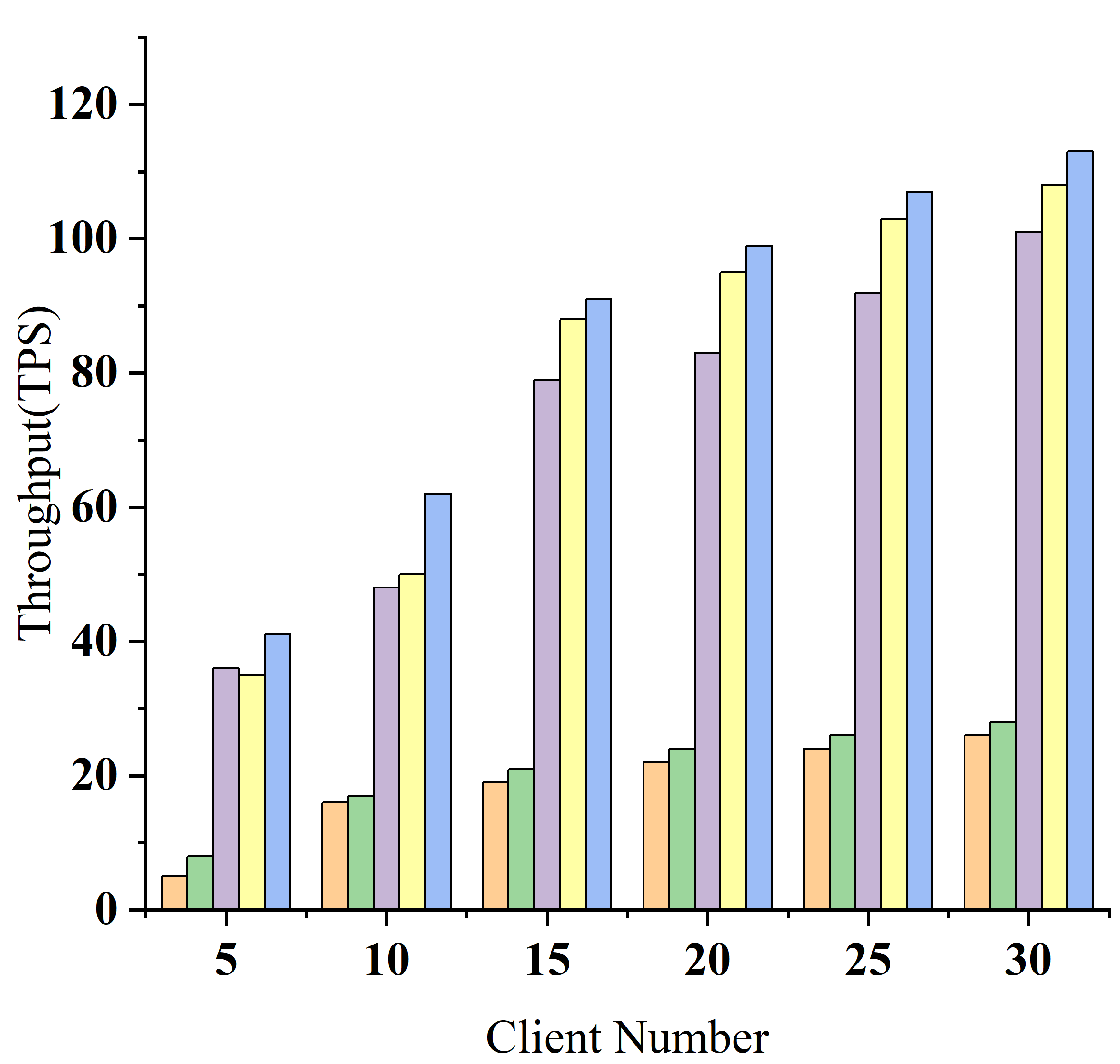}
         \caption{}
        \label{fig:subfig2}
    \end{subfigure}
    \begin{subfigure}[b]{0.22\textwidth}
        \centering
        \includegraphics[width=\textwidth]{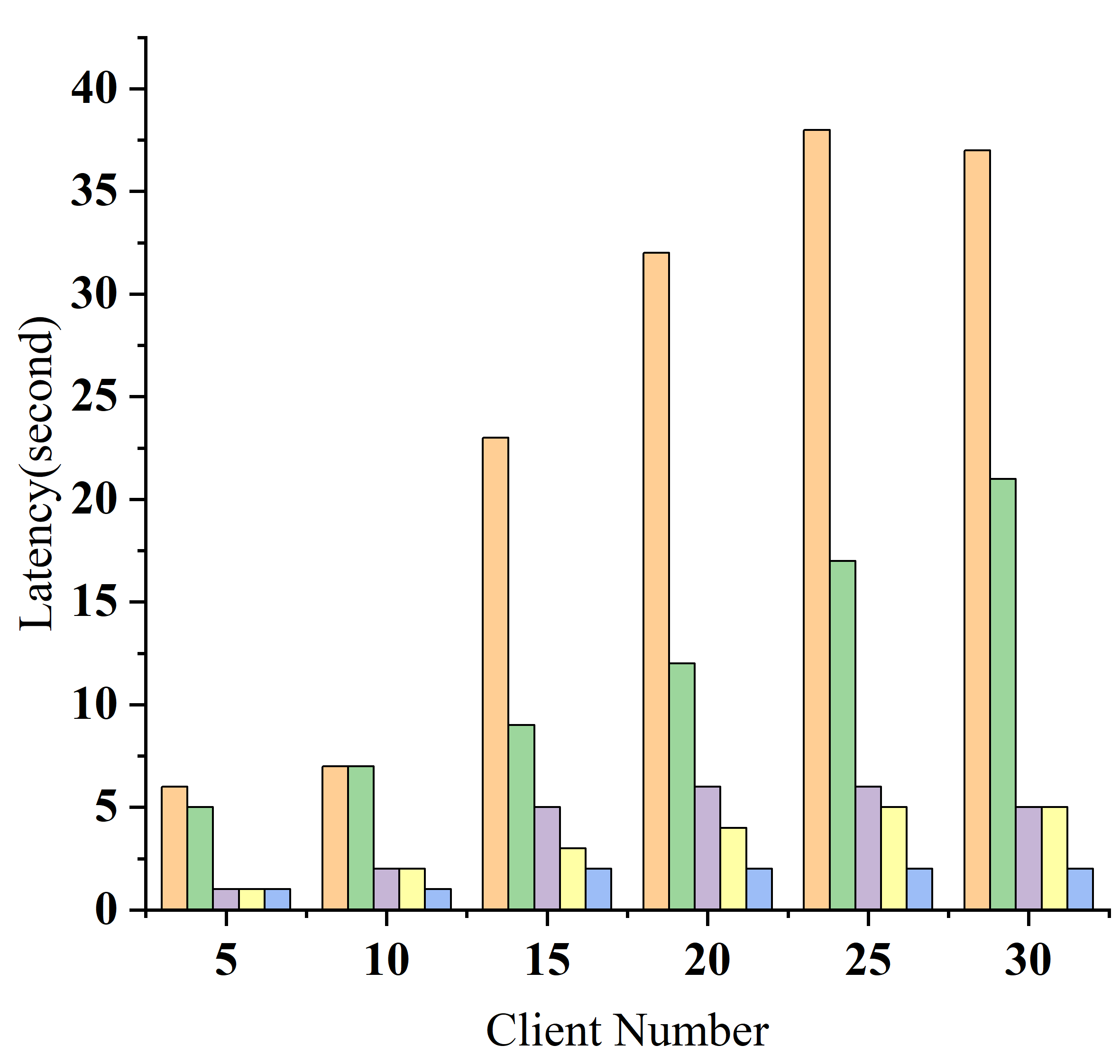}
         \caption{}
        \label{fig:subfig3}
    \end{subfigure}
    \hfill
    \begin{subfigure}[b]{0.22\textwidth}
        \centering
        \includegraphics[width=\textwidth]{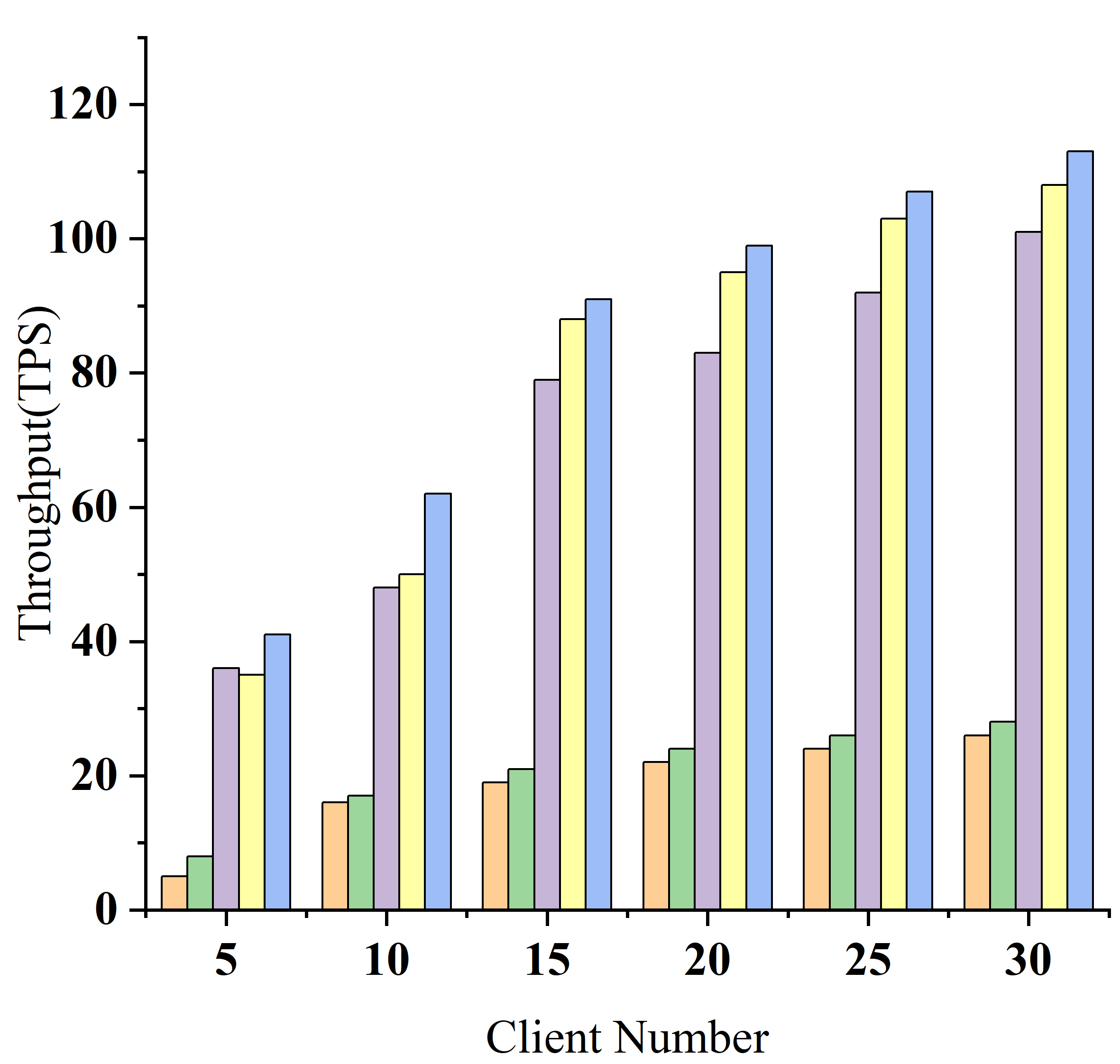}
         \caption{}
        \label{fig:subfig4}
    \end{subfigure}
    \caption{Performance Comparison.(a) Throughput for uploading updated models, (b) Throughput for querying global model or tip nodes, (c) Latency for uploading updated models, (d) Latency for querying.}
    \label{fig:3images}
\end{figure}
\vspace{-8pt}

Fig.~\ref{fig:3images} illustrates the throughput and latency results for CIFAR-10. As shown, \methodname{} achieves 56 TPS for uploading models and 120 TPS for querying at 30 clients, with a latency of 4 seconds for both tasks. Its advantage lies in uploading only metadata, reducing communication overhead. In contrast, BlockFL and BFLC, with throughputs of 20 TPS and 25 TPS and latencies over 28 seconds, suffer from bandwidth bottlenecks. ScaleSFL and DAG-FL improve request handling but still fall short, with querying throughputs of 95 TPS and 110 TPS, respectively.
\section{conclusion}
In this paper, we present a DAG-based asynchronous federated learning framework, namely \methodname{}. We introduce a tip selection method that considers time freshness, node reachability, and model accuracy, thereby refining model updates and better aligning them with recent training progress. A DAG verification mechanism prevents task tampering, ensuring data trustworthiness. Experiments on multiple benchmark datasets demonstrate that \methodname{} outperforms existing methods in training efficiency, model accuracy, throughput, and latency.
\section*{Acknowledgment}
This research is supported, in part, by the National Key R\&D Program of China (NO.2021YFF0704102) and the Natural Science Foundation of China(No. 92367202), the major Science and Technology Innovation of Shandong Province (2024CXGC010101), the Youth Student Fundamental Research Funds of Shandong University. This research is also supported, in part, by the Ministry of Education, Singapore, under its Academic Research Fund Tier 1; the National Research Foundation, Singapore and DSO National Laboratories under the AI Singapore Programme (AISG Award No. AISG2-RP-2020-019).

\bibliographystyle{IEEEbib}
\bibliography{icme2025references}

\end{document}